\pdfoutput=1

\documentclass[11pt]{article}

\usepackage{acl}

\usepackage{times}
\usepackage{latexsym}
\usepackage{longtable}

\usepackage[T1]{fontenc}

\usepackage[utf8]{inputenc}

\usepackage{microtype}

\usepackage{inconsolata}

\usepackage{graphicx}

\usepackage{xspace}
\usepackage{booktabs}
\usepackage{multirow}
\usepackage{algorithm}
\usepackage{algpseudocode}
\usepackage{xcolor}
\usepackage[mathletters]{ucs}
\usepackage{supertabular}
\usepackage{multicol}
\title{A Case Study of Cross-Lingual Zero-Shot Generalization for\\ Classical Languages in LLMs}

\author{
V.S.D.S. Mahesh Akavarapu$^{\alpha}$,
Hrishikesh Terdalkar$^{\beta}$,
Pramit Bhattacharyya$^{\gamma}$,\\
\bf
Shubhangi Agarwal$^{\beta}$,
Vishakha Deulgaonkar$^{\gamma}$,
Pralay Manna$^{\gamma}$,
Chaitali Dangarikar$^{\gamma}$,\\
\bf
Arnab Bhattacharya$^{\gamma}$ \\
$^{\alpha}$University of Tübingen, $^{\beta}$University of Lyon 1, $^{\gamma}$Indian Institute of Technology Kanpur\\
\texttt{mahesh.akavarapu@uni-tuebingen.de, hrishikesh.terdalkar@liris.cnrs.fr,}\\
\texttt{arnabb@cse.iitk.ac.in}}

\newcommand{\ab}[1]{\textcolor{red}{AB: #1}}

\newcommand{\draft}[1]{\textcolor{gray}{{\it #1}}}

\newcommand{\ramayana}{R\={a}m\={a}ya\d{n}a\xspace}
\newcommand{\bpn}{Bh\={a}vaprak\={a}\'{s}anigha\d{n}\d{t}u\xspace}
\newcommand{\ayurveda}{\={A}yurveda\xspace}

\newcommand{\countramayana}{1000\xspace}
\newcommand{\countayurveda}{2600\xspace}
\newcommand{\countmanual}{70\xspace}

\newcommand{\rmkgnodes}{867\xspace}
\newcommand{\rmkgrelations}{944\xspace}
\newcommand{\bpnkgnodes}{4685\xspace}
\newcommand{\bpnkgrelations}{10596\xspace}

\newcommand{\comment}[1]{}

\begin{document}

\maketitle

\begin{abstract}
Large Language Models (LLMs) have demonstrated remarkable generalization capabilities across diverse tasks and languages. In this study, we focus on natural language understanding in three classical languages---Sanskrit, Ancient Greek and Latin---to investigate the factors affecting cross-lingual zero-shot generalization. First, we explore named entity recognition and machine translation into English. While LLMs perform equal to or better than fine-tuned baselines on out-of-domain data, smaller models often struggle, especially with niche or abstract entity types. In addition, we concentrate on Sanskrit by presenting a factoid question–answering (QA) dataset and show that incorporating context via retrieval-augmented generation approach significantly boosts performance. In contrast, we observe pronounced performance drops for smaller LLMs across these QA tasks. These results suggest model scale as an important factor influencing cross-lingual generalization. Assuming that models used such as GPT-4o and Llama-3.1 are not instruction fine-tuned on classical languages, our findings provide insights into how LLMs may generalize on these languages and their consequent utility in classical studies.

\end{abstract}

\section{Introduction}
\label{sec:intro}
Large Language Models (LLMs) \citep{brown2020language, ouyang2022training, touvron2023llama} are known to generalize across various tasks using data from languages present in their pre-training phase, even when not present in instruction tuning \citep{wang-etal-2022-super, muennighoff-etal-2023-crosslingual, han-etal-2024-deep}. Previous work has demonstrated generalization to several low-resource languages via few-shot in-context learning \citep{cahyawijaya-etal-2024-llms}. In this study, we focus on zero-shot generalization to natural language understanding (NLU) tasks for three \emph{classical languages}---Sanskrit (\texttt{san}), Ancient Greek (\texttt{grc}), and Latin (\texttt{lat})---with a primary focus on Sanskrit. These languages represent a unique category of low-resource languages -- despite scarce data for downstream NLU tasks, they have rich ancient literature available in digitized formats \citep{goyal-etal-2012-distributed, berti2019digital}, and they exert significant influence on the vocabulary and narrative structures of better-resourced languages (e.g., Latin contributes approximately 28\% of English vocabulary \citep{finkenstaedt1973ordered}). Moreover, the high inflection of these languages presents a challenge.

\begin{figure*}[t]
    \centering
    \includegraphics[width=\linewidth]{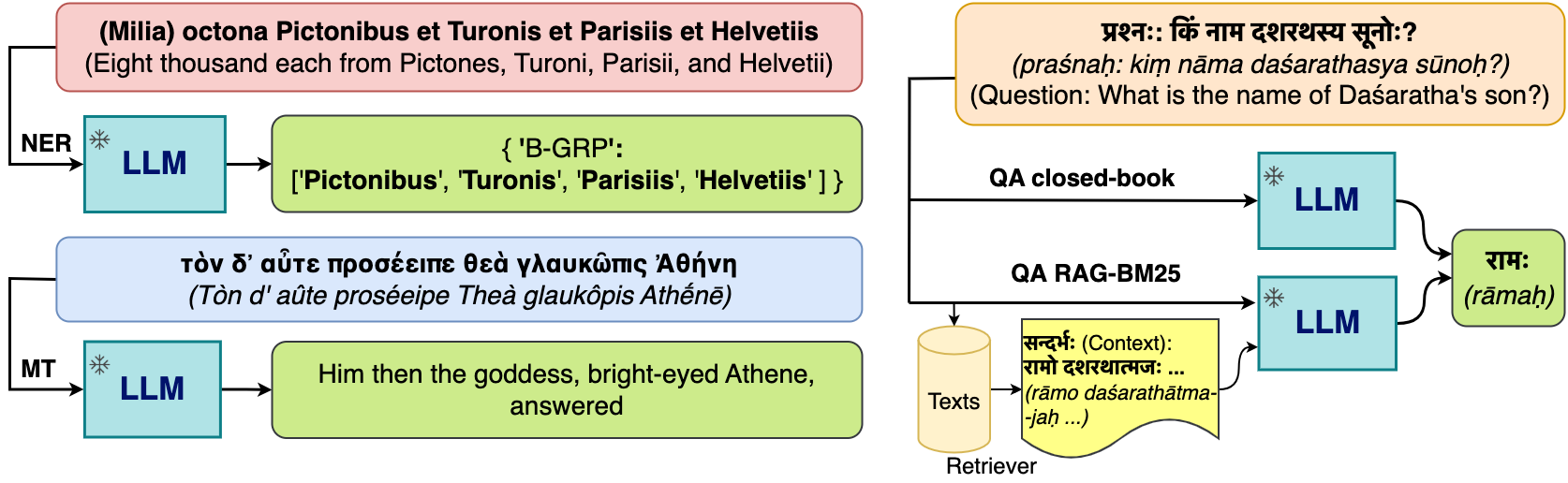}
    \caption{The three NLU tasks evaluated on the classical languages: Named-Entity Recognition (top-left), Machine Translation (bottom-left) and Question-Answering (right).}
    \label{fig:sktqa_intro}
\end{figure*}

To investigate these issues, we conduct two sets of \emph{zero-shot} experiments using \texttt{gpt-4o} \citep{openai2024gpt4o, OpenAI_GPT4_2023}, \texttt{llama-3.1-405b-instruct} \citep{dubey2024llama3herdmodels}, and their smaller variants. First, we assess performance on two NLU tasks with available datasets for all three languages, namely, named entity recognition (NER) and machine translation to English (MT). We observe that larger models perform comparably or even better than previously reported fine-tuned models on out-of-domain data. Second, we focus on Sanskrit by introducing a factoid closed-book QA dataset and show that question-answering performance improves with retrieval augmented generation (RAG) \citep{lewis2020retrieval} when the model is provided with relevant texts. The tasks are illustrated in Figure~\ref{fig:sktqa_intro}.

Given the recent nature of these datasets relative to the models' knowledge cut-off dates, and the likelihood that instruction-tuning on these languages is limited, the robust performance observed can be attributed to cross-lingual generalization. We refer to our prompting strategy as zero-shot, as it includes no task-specific examples, and it is unlikely that such examples in these languages were present in the models' training or instruction-tuning data. In both experimental setups, we find that smaller models struggle, particularly with niche entity types in NER, and in effectively leveraging contextual information via RAG, thereby suggesting model scale as an important factor.

\section{Datasets and Methods}
\label{sec:datasets}

\begin{table}[t]
\resizebox{\columnwidth}{!}{
\begin{tabular}{lclr} \toprule
\textbf{Task}        & \textbf{Language} & \textbf{Source}                                      & \textbf{Size} \\ \midrule
\multirow{3}{*}{NER} & \texttt{san}               & \citet{terdalkar2024sanskrit}        & 139                                \\
                     & \texttt{lat}              & \citet{erdmann-etal-2019-practical}  & 3410                               \\
                     & \texttt{grc}              & \citet{myerston-nereus-2025}         & 4957                               \\ \midrule
\multirow{3}{*}{MT-\texttt{en}}  & \texttt{san}               & \citet{maheshwari-etal-2024-samayik} & 6464                               \\
                     & \texttt{lat}              & \citet{rosenthal-machina-2023}       & 1014                               \\
                     & \texttt{grc}               & \citet{Palladino-2023}               & 274                                \\ \midrule
QA                   & \texttt{san}               & Ours                                                  & 1501                               \\
\bottomrule
\end{tabular}
}
\caption{An overview of NLU datasets used in this study for Sanskrit (\texttt{san}), Latin (\texttt{lat}) and Ancient Greek (\texttt{grc}). QA dataset for \texttt{san} is contributed in this work. Size represents the number of sentences of test sets (wherever train-test splits exist).}
\label{tab:datasets}
\end{table}

The datasets used in our experiments are summarized in Table \ref{tab:datasets}. Our aim is to evaluate zero-shot capabilities where evaluation is done directly on test data without fine-tuning on the training data. Thus, we only consider the test sets of these datasets. Notably, the Sanskrit NER dataset (\texttt{san}) is the smallest, comprising 139 sentences (1558 tokens) \citep{terdalkar2024sanskrit}. In addition to these publicly available datasets, we contribute a new factoid closed-domain QA dataset in Sanskrit, described in detail in Section \ref{subsec:sanqa}.

We evaluate the zero-shot capabilities of both large and small variants of our models: proprietary (\texttt{gpt-4o} and \texttt{gpt-4o-mini} \citep{openai2024gpt4o}) and open-source (\texttt{llama-3.1-405b-instruct} and \texttt{llama-3.1-8b-instruct} \citep{dubey2024llama3herdmodels}). According to official documentation, these models have knowledge cut-off dates at the end of 2023. Many datasets considered in this work (Table \ref{tab:datasets}) are released beyond these dates, in other words, they are unlikely to be seen in the pre-training data of these models. Given that none of the documentation indicates explicit instruction tuning on Sanskrit, Ancient Greek, or Latin, any observed performance in these languages is likely attributable to cross-lingual generalization. Previous zero-shot applications of LLMs to classical languages have been limited to Latin machine translation and summarization \citep{volk-etal-2024-llm}, although several works have been dedicated to building language models for these languages \citep{riemenschneider-frank-2023-exploring, nehrdich-etal-2024-one}, however, with fine-tuning restricted to morphological parsing-related tasks like dependency parsing \citep{nehrdich-hellwig-2022-accurate, hellwig2023data, sandhan-etal-2023-systematic}. 

All prompts used for these tasks are provided in Appendix~\ref{sec:app-prompts}. The prompts are tested in both English and the respective languages.

\subsection{Sanskrit QA}
\label{subsec:sanqa}

To further evaluate comprehension, we focus on question-answering (QA) in Sanskrit -- a domain with very limited datasets. The only existing dataset by \citet{terdalkar-bhattacharya-2019-framework} comprises 80 kinship-related questions. To address this gap, we created a new dataset containing 1501 factoid QA pairs covering distinct domains in Sanskrit: epics and healthcare. Specifically, we selected two key texts: (1) \ramayana, an ancient Indian epic, and (2) \bpn, a foundational text on \ayurveda. Details of the dataset are provided in Appendix~\ref{sec:app-qadata}.

\begin{figure}[t]
    \centering
    \includegraphics[width=0.48\linewidth]{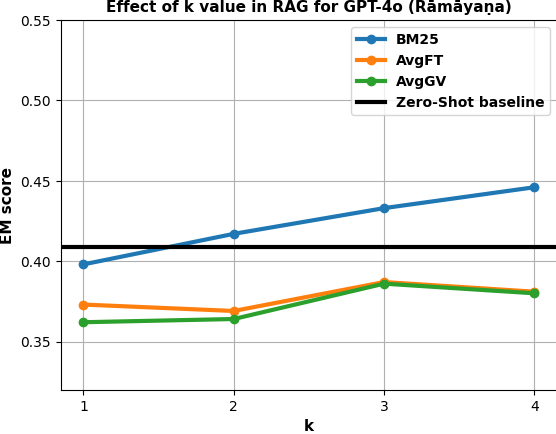}
    \includegraphics[width=0.48\linewidth]{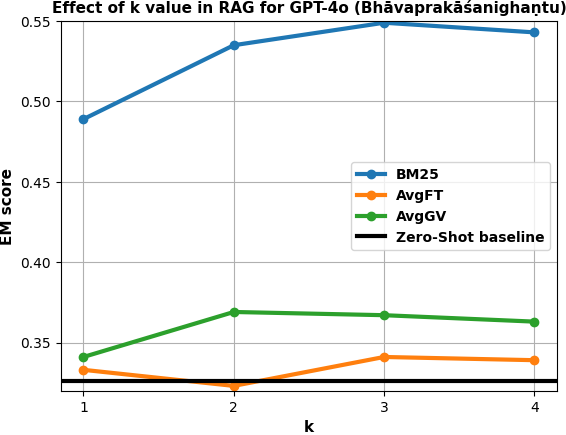}
    \caption{Effect of $k$ on RAG, denoting the number of top best matching text chunks retrieved, on the performances of GPT-4o with retrievers based on BM25, averaged FastText (AvgFT) and GloVe (AvgGV) embeddings respectively of datasets \ramayana (left) and \bpn (right).}
    \label{fig:ragk}
\end{figure}

For QA evaluation, we employ a closed-book setting using prompts detailed in Appendix~\ref{subsec:app-prompt-zs}. To emulate extractive QA, we implement a Retrieval-Augmented Generation (RAG) approach by retrieving the top-$k$ relevant passages ($k$ tuned to 4) from the original Sanskrit texts using BM25 \citep{sparck1972statistical, robertson2009probabilistic}. We also compare BM25 with embedding-based retrievers—FastText \citep{bojanowski2017enriching} and GloVe \citep{pennington-etal-2014-glove}—and vary $k$ to assess performance using \texttt{gpt-4o} with Sanskrit prompts. As shown in Fig.~\ref{fig:ragk}, BM25 consistently outperforms the embedding-based methods, and $k = 4$ emerges as an optimal choice across metrics.

To examine whether the inclusion of answer-bearing contexts benefits model performance, we manually annotated the relevance of retrieved passages. Since BM25 relies on lexical similarity, typically favoring lemmas over inflected forms, we introduce a lemmatization step using a transformer-based Seq2Seq Sanskrit lemmatizer trained on the DCS corpus \citep{hellwig2010dcs}, achieving a mean F1 score of 0.94 on a held-out test set. Further details on RAG and lemmatization are provided in Appendix~\ref{sec:rag}, and implementation details in Appendix~\ref{sec:app-impl}. Code and data are available at \url{https://github.com/mahesh-ak/SktQA}.

\section{Results}
\label{sec:eval}

\begin{figure*}[t]
    \centering
    \includegraphics[width=\textwidth]{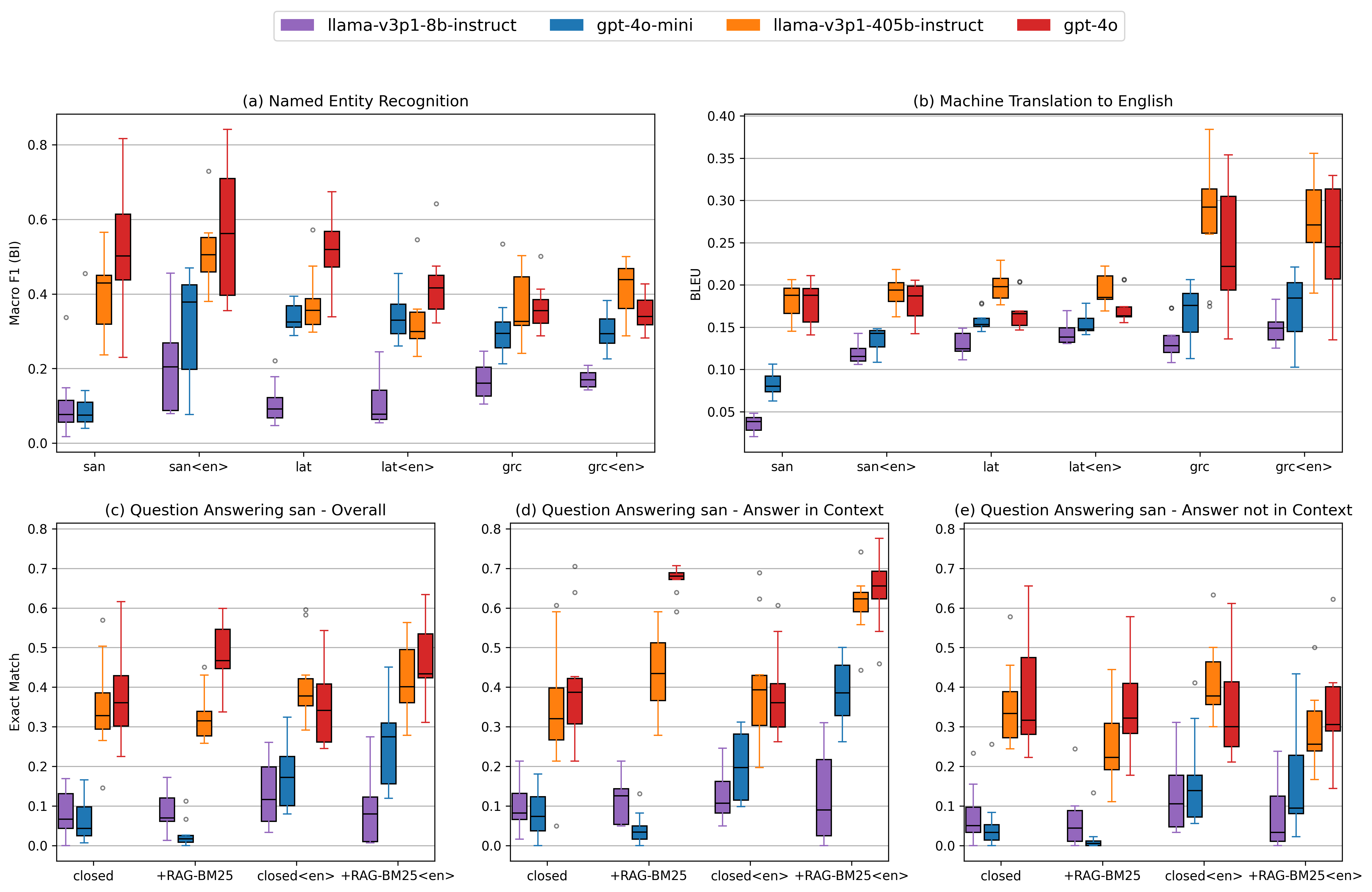}
    \caption{Zero-shot evaluation of LLMs on three NLU tasks for classical languages (language codes in ISO 639-2). Prompts when in English are denoted by \texttt{<en>}, otherwise are in respective languages. Larger LLMs are represented in {\color{red} red} and {\color{orange} orange}, while smaller LLMs in {\color{teal} blue} and {\color{violet} purple}. First row displays the performances on NER (a) and MT (to \texttt{en}) (b) for all three languages. Second row displays the performances on QA for Sanskrit alone. Out of 1501 QA pairs considered (c), 607 QA pairs are with answer present in at least one of the $k=4$ contexts extracted by BM25 and 894 QA pairs with answer not inferable from contexts, which are respectively considered in (d) and (e).}
    \label{fig:results}
\end{figure*}

Figure~\ref{fig:results} presents our zero-shot evaluation results, demonstrating that larger LLMs exhibit robust cross-lingual generalization across four NLU tasks---named entity recognition (NER), machine translation (MT), closed-book QA, and extractive QA (simulated via RAG-BM25)---in three classical languages (with QA evaluated solely on Sanskrit). To assess variability, each test set is segmented into 10 chunks during evaluation resulting in a box-plot. Larger models perform better than previous fine-tuned models on out-of-domain data as reported in Appendix~\ref{sec:app-add}. Notably, when answer-bearing contexts are provided (Fig.~\ref{fig:results}d) versus when they are absent (Fig.~\ref{fig:results}e), the models show significant performance gains, suggesting comprehension abilities in Sanskrit, a language characterized by high inflection. This behavior is however, not exhibited by smaller models when prompted in Sanskrit.

\subsection{Prompt Language: English versus Native}

During evaluation, we prompted models both in English and in each target language. As shown in Figure~\ref{fig:results}, English prompts generally outperform Sanskrit prompts, particularly with smaller models, providing partial evidence that these models have not been instruction-tuned on Sanskrit \citep{muennighoff-etal-2023-crosslingual}. For Latin and Ancient Greek, this English‑prompt advantage holds mainly for smaller models; larger models perform equally well, or even better, with native-language prompts (e.g., in Latin NER). This does not imply instruction tuning in these languages, since larger and smaller models likely saw comparable amounts of tuning data. Rather, it suggests that cross-lingual transfer is especially strong for Latin and Ancient Greek in larger models, reflecting their substantial influence on high-resource languages such as English.

\begin{figure*}[t]
    \includegraphics[width=\textwidth]{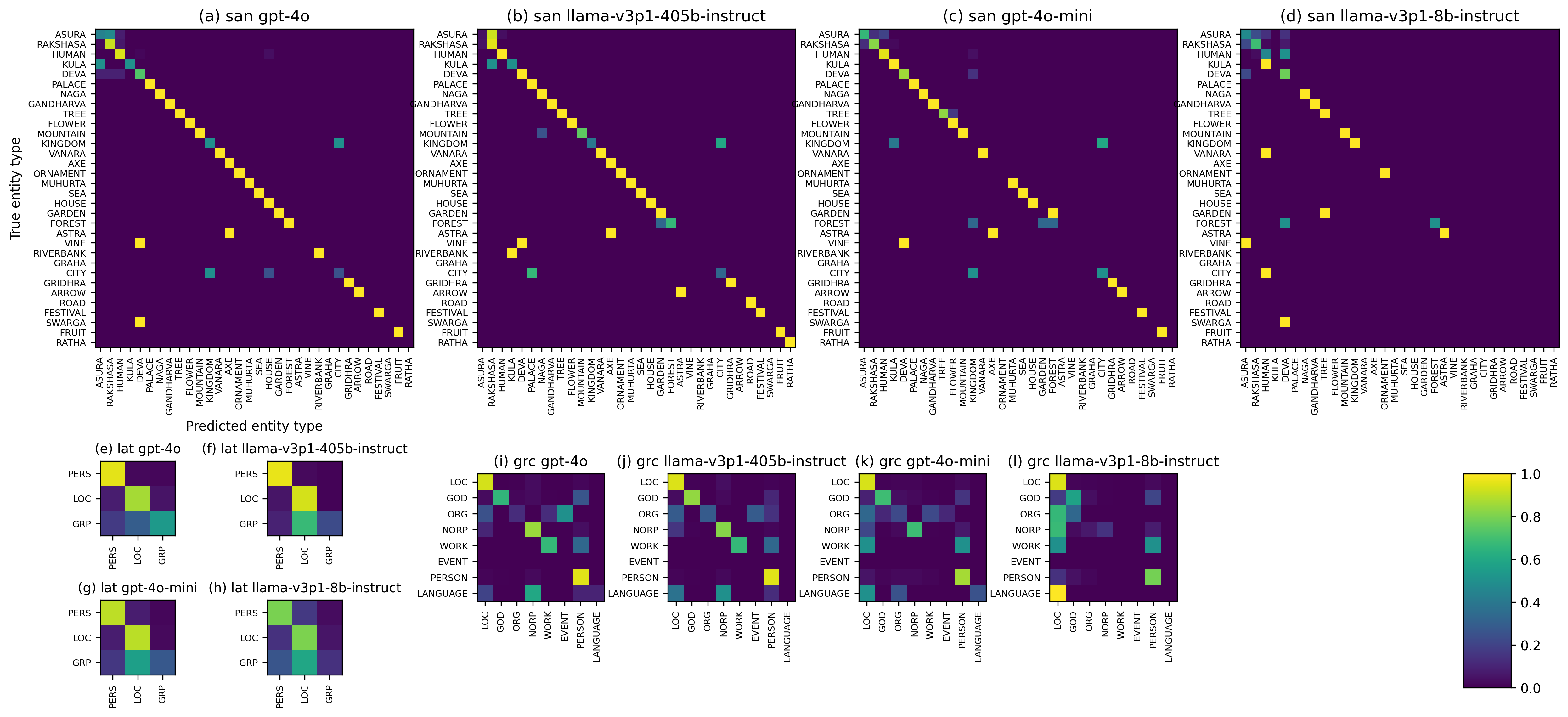}
    \caption{Confusion matrices from the NER task in \texttt{san} (a-d), \texttt{lat} (e-h) and \texttt{grc} (i-l), all with \texttt{<en>} prompts, normalized across rows.}
    \label{fig:ner_cm}
\end{figure*}

\subsection{Misclassified Entities in NER}

Figure~\ref{fig:ner_cm} displays confusion matrices for the NER task. Across all three languages, the smaller models exhibit more confusion among semantically related classes (see \ref{app:nertypes} for descriptions of entity types), while the larger models show fewer off‐diagonal errors. In \texttt{san}, mythological entities like Deva, Asura, and Rakshasa or semantically closed entities like Kingdom versus City (e.g., Ko\'{s}ala vs Ayodhy\={a}) or Forest (e.g., Da\d{n}\d{d}aka) versus Garden (e.g., Nandana) often get misclassified with each other in the smaller models. For \texttt{lat}, entity type GRP proves troublesome for the smaller models, suggesting struggles in separating individual entities from collective ones. In \texttt{grc}, categories such as LOC and ORG show higher confusion counts akin to GRP in \texttt{lat} while confusion between God and Persons seems similar to what was discussed for Sanskrit. In contrast, much clearer boundaries emerge in the larger models' confusion matrices. In many of these cases, the domain or style of the texts, especially if they involve mythological or archaic terms typical of classical texts, can be understood to influence performance, as models with less exposure to specialized terminology may conflate related entity types.

\subsection{Inflection in Answers in Sanskrit QA}

\begin{table}[t]
\resizebox{\columnwidth}{!}{
\begin{tabular}{l|rr|rr} \toprule
\multirow{2}{*}{\textbf{LLM}} & \multicolumn{2}{c|}{\textbf{Closed-book}}                       & \multicolumn{2}{c}{\textbf{+ RAG-BM25}}                        \\
\cmidrule{2-3}
\cmidrule{4-5}
                              & \multicolumn{1}{l}{Inflected} & \multicolumn{1}{l|}{Lemmatized} & \multicolumn{1}{l}{Inflected} & \multicolumn{1}{l}{Lemmatized} \\ \midrule
gpt-4o                        & 0.36                          & 0.37                           & 0.46                          & 0.48                           \\
llama-3.1-405b-instruct       & 0.41                          & 0.40                           & 0.42                          & 0.44                           \\
gpt-4o-mini                   & 0.18                          & 0.20                           & 0.25                          & 0.28                           \\
llama-3.1-8b-instruct         & 0.13                          & 0.15                           & 0.09                          & 0.10        \\ \bottomrule                  
\end{tabular}
}
\caption{Comparison of EM scores in \texttt{san} QA (English prompts) when predicted and gold answers are considered with inflection or lemmatized.}
\label{tab:inflc}
\end{table}

In the Sanskrit question-answering task, models are expected to generate single-word answers with the correct inflection. For computing exact match (EM) scores, we manually identified all acceptable answers, excluding those with incorrect inflection (e.g., wrong case or gender endings). To quantify inflection errors, we also calculated EM scores on lemmatized versions of the gold standard and predicted answers, as shown in Table~\ref{tab:inflc}. Most models show only a slight increase in EM scores on lemmatized answers, suggesting that inflection errors are relatively minor, a finding corroborated by manual inspection. Future work could extend this analysis to investigate inflection accuracy in full sentence generation within broader natural language generation scenarios.

\subsection{Sanskrit Orthography: Devanagari versus IAST}

\begin{table}[t]
\resizebox{\columnwidth}{!}{
\begin{tabular}{l|rr|rr} \toprule
\multirow{2}{*}{\textbf{LLM}} & \multicolumn{2}{c|}{\textbf{MT (BLEU)}}                       & \multicolumn{2}{c}{\textbf{NER (Macro F1-BI)}}                        \\
\cmidrule{2-3}
\cmidrule{4-5}
                              & \multicolumn{1}{l}{Devanagari} & \multicolumn{1}{l|}{IAST} & \multicolumn{1}{l}{Devanagari} & \multicolumn{1}{r}{IAST} \\ \midrule
gpt-4o                   & 0.179                                                           & 0.165                                                            & 0.637                                                             & 0.599                                                              \\
llama-v3p1-405b-instruct & 0.193                                                           & 0.148                                                            & 0.561                                                             & 0.556                                                              \\
gpt-4o-mini              & 0.135                                                             & 0.099                                                             & 0.359                                                             & 0.318                                                              \\
llama-v3p1-8b-instruct   & 0.120                                                           & 0.063                                                             & 0.164                                                             & 0.149     \\ \bottomrule                                                        
\end{tabular} 
}
\caption{Comparison of performances in \texttt{san} MT and NER (English prompts) when the input sentences are given Devanagari script or in IAST script.}
\label{tab:ortho}
\end{table}

So far, we have shown robust cross-lingual generalization in the models. We now turn to one possible underlying mechanism---orthographic transfer---where models benefit from shared scripts across languages. Prior work has identified orthography as a key factor in cross-lingual transfer for LLMs \citep{muller-etal-2021-unseen, fujinuma-etal-2022-match}. To isolate this effect, we re‑ran our Sanskrit NER and MT experiments (using English prompts) in Roman‑based IAST transliteration instead of Devanagari. Table \ref{tab:ortho} compares performance in both scripts. Models perform better with the Devanagari script, which is shared by higher‑resource relatives like Hindi and Marathi, reinforcing the importance of script sharing. However, results in IAST are only slightly lower, suggesting that Roman‑based transliterations also feature prominently in the pre‑training data. In future, we will investigate whether model outputs are consistent across both scripts, that is, whether these LLMs are effectively \textit{digraphic} in Sanskrit. 

\subsection{Knowledge-Graph Question-Answering}

Additionally, we explore the use of knowledge graphs (KGs) for Sanskrit QA. We evaluated a KG derived from the \bpn text \citep{terdalkar-etal-2023-semantic} and constructed a small KG for \ramayana (details in Appendix~\ref{subsec:kgdesc}). Using the Think-On-Graph (ToG) paradigm \citep{sun2024thinkongraph}, which iteratively explores the KG paths for answer retrieval in a training-free zero-shot manner \citep{xu-etal-2024-generate}, we observed that \texttt{gpt-4o} could effectively execute this method. Although it occasionally extracted correct answers, its performance did not significantly exceed that of the closed-book setting, most likely due to the incompleteness of the KGs (see \S\ref{subsec:llmkgres}). Future work may focus on developing more comprehensive KGs to enhance this retrieval method.
\section{Conclusions}
\label{sec:conc}

In summary, our zero-shot evaluations demonstrate that larger language models exhibit robust cross-lingual generalization across diverse natural language understanding tasks in classical languages, including NER, machine translation, and QA. Notably, the significant performance gains achieved when answer-bearing contexts are provided, particularly in Sanskrit QA, suggest comprehension abilities in highly inflected languages. Moreover, our contribution of a novel Sanskrit QA dataset provides a valuable resource for evaluating and benchmarking LLM performance on classical language tasks. Importantly, these models have not been explicitly instruction tuned on Sanskrit, Latin, or Ancient Greek---evidenced by the superior performance achieved when using English prompts for Sanskrit---which indicates that their zero-shot performance is attributable solely to cross-lingual generalization. 

Future work will focus on expanding dataset coverage, knowledge graphs and exploring additional classical languages and tasks, further advancing our understanding of cross-lingual generalization in LLMs and its applications in digital humanities and multilingual NLP research.


\newpage

\section*{Acknowledgements}

This research is financially supported by the Indian Knowledge Systems (IKS) Division of Ministry of Education, Govt.\ of India (project number AICTE/IKS/RFP1/2021-22/12). Mahesh Akavarapu received funding from Volkswagen Foundation under the Phylomilia project within the Pioneering Projects funding line. We also thank anonymous reviewers and the Area Chairs for their comments that have helped improve the paper.


\section*{Limitations}

While our study demonstrates robust cross-lingual generalization in large language models for classical languages, several limitations warrant discussion. First, our newly contributed Sanskrit QA dataset, although valuable, is limited in size. Our evaluation relies exclusively on zero-shot performance, as the models have not been explicitly instruction tuned on these languages; this design choice may obscure potential benefits achievable through targeted fine-tuning. Further, a few datasets we experimented were released within the models' knowledge cut-off dates raising the issue of data contamination. Among these, only Ancient Greek MT exhibits anomalously high performance, suggesting possible exposure. In general, NER, owing to its structural data should be less susceptible to contamination than MT. Furthermore, the effectiveness of our BM25-based retrieval approach depends heavily on preprocessing steps such as lemmatization, which might not optimally address all linguistic variations in highly inflected languages. Finally, our comparisons are based on a limited set of proprietary and open-source models, and future work should extend this analysis to a broader range of models and tasks to fully understand the nuances of cross-lingual generalization in classical languages.

\section*{Ethics Statement}
\label{sec:ethics}


Classical Sanskrit epics hold deep cultural and religious significance in Indian traditions, and similarly, \ayurveda represents a revered tradition-bound area within healthcare. We acknowledge that any research involving these subjects must be conducted with particular care. It is essential to note that, as with conventional treatment, \={A}yurvedic practices require professional consultation and should not be substituted by automated responses. Although our experiments indicate that paradigms like RAG produce more grounded and, hence, potentially safer outputs, there is no assurance that the responses from current LLMs in these domains meet clinical or religious safety standards. Consequently, the authors do not endorse using the datasets beyond the scope of linguistic research. These datasets are released for open-source, non-commercial use, and all annotators have been compensated at fair, standard rates.

\comment{
We have used \draft{publicly} available version of \ramayana and \bpn for the curation of KG and Question-Answering dataset. Hence, there has been no copyright infringement for the curation of KG and Question-Answering dataset. Ramayana being an Indian epic and Bhāvaprakāśanighaṇṭu being an Ayurvedic text are mostly devoid of any abusive statements. Hence, we have not focussed on seggregating any abusive statements from the text. 
All the annotators used for the project are paid at the standard rate of Rs 350 per hour. We will release all the data and codes associated with the paper upon acceptance of the paper under a non-commercial license.

\ab{We took the questions from this book \cite{ramayanabook,bpnbook} for reference, and framed Sanskrit questions ourselves.}
}


\bibliography{anthology,custom}

\newpage
\onecolumn
\appendix
\section*{Appendix}

\section{Prompts}
\label{sec:app-prompts}

The Sanskrit prompts are in Devanagari script. In this appendix, we provide these prompts transliterated in IAST scheme.

\subsection{Prompts for Named Entity Recognition}
\subsubsection*{Prompt in English}
Recognize the named entities from the following sentence in \{\texttt{LANGUAGE}\}. The valid tags are \{\texttt{ENTITY TYPES}\}. Do not provide explanation and do not list out entries of `O'. Example:\\
Sentence: <word\_1> <word\_2> <word\_3> <word\_4> <word\_5>\\
Output: \{\{`B-<entity1>': [`<word\_1>', `<word\_4>'], `B-<entity2>':[`<word\_5>']\}\}\\
Sentence: \{\texttt{INPUT}\}\\
Output:

(The example is never a real sentence and is only provided to specify the output structure. Hence, the evaluations are strictly zero-shot.)

\subsubsection*{Prompt in Sanskrit}
adho datta v\={a}kye n\={a}mak\d{r}t\={a}\d{h} sattv\={a}\d{h} (named entities) abhij\={a}n\={i}hi. tadapi viv\d{r}tam m\={a} kuru, kevalam p\d{r}\d{s}\d{t}a vi\d{s}ayasya uttaram dehi. api ca `O'-sambandhit\={a}ni na dey\={a}ni.\\
sattv\={a}\d{h} ete\d{s}u varge\d{s}u vartante - \{\texttt{ENTITY TYPES}\}. ud\={a}hara\d{n}\={a}ya - \\
v\={a}kyam: <padam\_1> <padam\_2> <padam\_3> <padam\_4> <padam\_5>\\
phalitam: \{\{ `B-<sattva\d{h}1>': [`<padam\_1>', `<padam\_4>'], `B-<sattva\d{h}2>': [`<padam\_5>']\}\}\\
v\={a}kyam: \{\texttt{INPUT}\} \\
phalitam:

\subsubsection*{Prompt in Latin}
Agnosce nomina propria (named entities) ex hac sententia Latina.
Notae validae sunt \{\texttt{ENTITY TYPES}\}.
Explanationem ne praebeas nec elementa `O' elenca. Exemplar:\\
Sententia: <verbum\_1> <verbum\_2> <verbum\_3> <verbum\_4> <verbum\_5>\\
Productus: \{\{`B-<entitatem1>': [`<verbum\_1>', `<verbum\_4>'], `I-<entitatem1>': [`<verbum\_2>'], `B-<entitatem3>':[`<verbum\_5>']\}\}\\
Sententia: \{\texttt{INPUT}\} \\
Productus:
           
\subsubsection*{Prompt in Ancient Greek}
$\mbox{'}A\nu\alpha\gamma\nu\acute{\omega}\rho\iota\sigma o\nu$ $\tau\grave{\alpha}$ $\mbox{'}o\nu \acute{o}\mu\alpha\tau\alpha$ (named entities) $\mbox{'}\varepsilon\kappa$ $\tau\tilde{\eta}\varsigma\delta\varepsilon$ $\tau\tilde{\eta}\varsigma$ $\mbox{`}E\lambda\lambda\eta\nu\iota\kappa\tilde{\eta}\varsigma$ $\pi\varepsilon\rho\iota \acute{o}\delta o\upsilon.$
$\tau\grave{\alpha}$ $\mbox{'}\acute{\varepsilon}\gamma\kappa\upsilon\rho\alpha$ $\varepsilon\mbox{'}\acute{\iota}\delta\eta$ $\mbox{'}\varepsilon\sigma\tau\iota\nu$ \{\texttt{ENTITY TYPES}\}.\\
NORP $\sigma\eta\mu\alpha\acute{\iota}\nu\varepsilon\iota$ $\mbox{'}\acute{\varepsilon}\theta\nu\eta$ ($o\tilde{\iota}o\nu$ $\mbox{`}\acute{E}\lambda\lambda\eta\nu\varepsilon\varsigma$, $\pi\acute{\varepsilon}\rho\sigma\alpha\iota$), $\mbox{'}\varepsilon\theta\nu\omega\nu\acute{\upsilon}\mu\iota\alpha$, $\kappa\alpha\grave{\iota}$ $\mbox{'}\acute{\alpha}\lambda\lambda\alpha\varsigma$ $\kappa o\iota\nu\omega\nu\iota\kappa\grave{\alpha}\varsigma$ $\mbox{`}o\mu\acute{\alpha}\delta\alpha\varsigma$ ($o\tilde{\iota}o\nu$ $\theta\rho\eta\sigma\kappa\varepsilon\upsilon\tau\iota\kappa\grave{\alpha}\varsigma$ $\mbox{`}o\rho\gamma\alpha\nu\acute{\omega}\sigma\varepsilon\iota\varsigma$).\\
$M\grave{\eta}$ $\pi\alpha\rho\acute{\varepsilon}\chi o\upsilon$ $\mbox{'}\varepsilon\xi\acute{\eta}\gamma\eta\sigma\iota\nu$ $\mbox{'}\varepsilon\nu$ \c{$\tau\tilde{\eta}$} $\mbox{'}\alpha\pi o\kappa\rho\acute{\iota}\sigma\varepsilon\iota$ $\mu\eta\delta\grave{\varepsilon}$ $\tau\grave{\alpha}$ $\varepsilon\grave{\iota}\varsigma$ `O' $\mbox{'}\varepsilon\gamma\gamma\varepsilon\gamma\rho\alpha\mu\mu\acute{\varepsilon}\nu\alpha$ $\pi\alpha\rho\alpha\tau\acute{\iota}\theta\varepsilon\sigma o$. $\pi\alpha\rho\acute{\alpha}\delta\varepsilon\iota\gamma\mu\alpha$:\\
$\pi\rho\acute{o}\tau\alpha\sigma\iota\varsigma$: <$\lambda\acute{\varepsilon}\xi\iota\varsigma$\_1> <$\lambda\acute{\varepsilon}\xi\iota\varsigma$\_2> <$\lambda\acute{\varepsilon}\xi\iota\varsigma$\_3> <$\lambda\acute{\varepsilon}\xi\iota\varsigma$\_4> <$\lambda\acute{\varepsilon}\xi\iota\varsigma$\_5>\\
$\pi\alpha\rho\alpha\gamma\omega\gamma\acute{\eta}$: \{\{`B-<$\mbox{'}O\nu\tau\acute{o}\tau\eta\varsigma$1>': [`<$\lambda\acute{\varepsilon}\xi\iota\varsigma$\_1>', `<$\lambda\acute{\varepsilon}\xi\iota\varsigma$\_4>'], `B-<$\mbox{'}O\nu\tau\acute{o}\tau\eta\varsigma$2>':['<$\lambda\acute{\varepsilon}\xi\iota\varsigma$\_5>']\}\}\\
$\pi\rho\acute{o}\tau\alpha\sigma\iota\varsigma$: \{\texttt{INPUT}\} \\
$\pi\alpha\rho\alpha\gamma\omega\gamma\acute{\eta}$:

\subsection{Prompts for Machine Translation}
\subsubsection*{Prompt in English}
Translate the following sentence in \{LANGUAGE\} into English. Do not give any explanations.

\subsubsection*{Prompt in Sanskrit}
adho datta-sa\d{m}sk\d{r}ta-v\={a}kyam \={a}\.{n}gle anuv\={a}daya, tad api viv\d{r}tam m\={a} kuru - 

\subsubsection*{Prompt in Latin}
Verte hanc sententiam Latinam in Anglicam. Nullam explicationem praebe.

\subsubsection*{Prompt in Ancient Greek}
$M\varepsilon\tau\acute{\alpha}\phi\rho\alpha\sigma o\nu$ $\tau\grave{\eta}\nu\delta\varepsilon$ $\tau\grave{\eta}\nu$ $\mbox{`}E\lambda\lambda\eta\nu\iota\kappa\grave{\eta}\nu$ $\pi\rho\acute{o}\tau\alpha\sigma\iota\nu$ $\varepsilon\grave{\iota}\varsigma$ $\tau\grave{\eta}\nu$ $\mbox{'}A\gamma\gamma\lambda\iota\kappa\acute{\eta}\nu$. $M\eta\delta\varepsilon\mu\acute{\iota}\alpha\nu$ $\mbox{'}\varepsilon\xi\acute{\eta}\gamma\eta\sigma\iota\nu$ $\pi\alpha\rho\acute{\varepsilon}\chi o\upsilon$.

\subsection*{(Sanskrit QA Prompts)}
In the following prompts, \texttt{TOPIC} is either `\ramayana' or `\ayurveda'.

\subsection{Prompts for Closed-book QA}
\subsubsection*{Prompt in English}
Answer the question related to \{\texttt{TOPIC}\} in the Sanskrit only. Give a single word answer if reasoning is not demanded in the answer. With regards to how-questions, answer in a short phrase, there is no single word restriction.\\
\{\texttt{QUESTION}\} \{\texttt{CHOICES}\} 

\subsubsection*{Prompt in Sanskrit}
\label{subsec:app-prompt-zs}
tvay\=a sa\d{m}sk\d{r}ta-bh\=a\d{s}\=ay\=am eva vaktavyam. na tu any\=asu bh\=a\d{s}\=asu. adha\d{h} \{\texttt{TOPIC}\}-sambandhe p\d{r}\d{s}\d{t}a-pra\'snasya pratyuttara\d{m} dehi. tadapi ekenaiva padena yadi uttare k\=ara\d{n}am n\=apek\d{s}itam. katham kimartham ity\=adi\d{s}u ekena laghu v\=akyena uttara\d{m} dehi atra tu eka-pada-niyama\d{h} n\=asti.\\
\{\texttt{QUESTION}\} \{\texttt{CHOICES}\}

\subsection{Prompts for RAG-QA}
\subsubsection*{Prompt in English}
Answer the following question related to \{\texttt{TOPIC}\} in Sanskrit only. Give a single word answer if reasoning is not demanded in the answer. With regards to how-questions, answer in a short phrase. Also take help from the contexts provided. The contexts may not always be relevant."

contexts: \{\texttt{CONTEXTS}\}

question:\{\texttt{QUESTION}\} \{\texttt{CHOICES}\} 

\subsubsection*{Prompt in Sanskrit}
\label{subsec:app-prompt-rag}
tvay\=a sa\d{m}sk\d{r}ta-bh\=a\d{s}\=ay\=am eva vaktavyam. na tu any\=asu bh\=a\d{s}\=asu. adha\d{h} \{\texttt{TOPIC}\}-sambandhe p\d{r}\d{s}\d{t}a-pra\'snasya pratyuttara\d{m} dehi. tadapi ekenaiva padena, y\={a}vad laghu \'sakya\d{m} t\=avad, ta\d{m} puna\d{h} viv\d{r}tam m\=a kuru. api ca yath\=a'va\'syam adha\d{h} datta-sandarbhebhya\d{h} ekatam\=at sah\=ayya\d{m} g\d{r}h\=a\d{n}a. tattu sarvad\=a s\=adhu iti n\=a'sti prat\=iti\d{h}.

sandarbh\=a\d{h}: \{\texttt{CONTEXTS}\}

pra\'sna\d{h}: \{\texttt{QUESTION}\} \{\texttt{CHOICES}\}

\section{Question Answering Dataset}
\label{sec:app-qadata}

In this appendix, we describe the creation of Sanskrit QA dataset.



We referred to two books that contain multiple-choice questions (MCQs) with answers: one
comprising \countramayana MCQs on \ramayana \cite{singh2009ramayana1000},
and another featuring a collection of \countayurveda questions from three
prominent texts of \ayurveda \cite{phull2017ayurvedamcq}. The
questions and options in these books are in Hindi language.

We carefully selected a relevant subset of questions from these books, including
all \countramayana questions from \ramayana dataset and $431$ from that of \ayurveda.
These questions are then translated into Sanskrit with the help of experts in
the language who are also familiar with the original Sanskrit texts. Further,
we consulted with a specialist in \ayurveda to review and discard incorrect
question-answer pairs, as well as to generate \countmanual new questions based
on \bpn. Ultimately, the question-answering dataset consists of $1501$
questions. 

The answers typically agree in grammatical case with the corresponding interrogative of the
question. The following is a question-answer pair as an
illustration\footnote{\textbf{gen} - genitive, \textbf{loc} - locative}:

\begin{table}[h!]

\begin{tabular}{lllllll}
\textbf{Q:} \textit{\'s\=itala-jalasya} & \textit{p\=ana\d{m}} & \textit{kasmin} &  \textit{roge} & \textit{ni\d{s}iddham} & \textit{asti}? & \textbf{A:} \textit{gala-grahe}\\
\textbf{Q:} cold-water.\textbf{gen} & drinking & what.\textbf{loc} &  disease.\textbf{loc} & forbidden & is & \textbf{A:} pharyngitis.\textbf{loc} \\
\end{tabular}
\caption*{Question: During which condition is the drinking of cold water forbidden? Answer: During pharyngitis.}
\end{table}

Most questions in the datasets have a single-word answer except a few including those in the \ramayana that fall under the category `Origins' (Table~\ref{tab:catramayana}). An example question-answer pair under this category that demands reasoning in the answer:

\textbf{Q:} \textit{r\=aj\=a-sagare\d{n}a sagara\d{h} iti n\=ama kuta\d{h} pr\=aptam?}

``How did King Sagara obtain such a name?''

\textbf{A:} \textit{saha tena gare\d{n}aiva j\=ata\d{h} sa sagaro 'bhavat}

``He was indeed born along with (\textit{sa-}) the poison (\textit{gara}), thus he became \textit{Sagara}.''

For such questions (only about 50), the answers can be paraphrased variously, thereby requiring manual evaluation.


The broad semantic and domain-specific categories of the questions are detailed in Tables~\ref{tab:catramayana}~and~\ref{tab:catayurveda}.
\begin{table}[t]
    \begin{minipage}{.5\linewidth}
      \centering
        \resizebox{\textwidth}{!}{
    \begin{tabular}{lp{0.72\textwidth}r}
        \toprule
        Category & Description & \#Q \\
        \midrule
        Names & Names of various characters & 97\\
        Actions & Who performed certain actions? & 47\\
        Origins & Origin of various names & 49\\
        Numeric & Questions with numerical answers & 79\\
        Quotes & Who said to whom? & 31\\
        Boons and Curses & Who endowed boons / curses on whom & 31\\
        Weapons & Questions related to various types of weapons & 59\\
        Locations & Locations of important events or characters & 71\\
        Kinship & Questions pertaining to human kinship relationships & 133\\
        Slay & Who slayed whom & 49\\
        Kingdoms & Which king ruled which kingdom & 27\\
        Incarnations & Who were incarnations of which deities & 27\\
        MCQ & Multiple choice questions & 140\\
        Miscellaneous & Other questions & 196\\
        \bottomrule
    \end{tabular}
    }
    \caption{Question Categories for \ramayana QA Dataset}
    \label{tab:catramayana}
    \end{minipage}%
    \begin{minipage}{.5\linewidth}
      \centering
        \resizebox{\textwidth}{!}{
    \begin{tabular}{lp{0.85\textwidth}r}
        \toprule
        Category & Description & \#Q \\
        \midrule
        Synonym & Synonyms of substances & 174\\
        Type & Variants or types of substances & 30\\
        Property & Properties of substances & 20\\
        Comparison & Comparison between properties of various substances or their variants & 24\\
        Consumption & Related to consumption of medicine including suitability, method, accompaniments etc. & 23\\
        Count & Counting types or properties of substances & 59\\
        Quantity & Quantity of substances in various procedures or methods & 21\\
        Time-Location & Time or location in the context of substances or methods & 17\\
        Effect & Effect of substances & 15\\
        Treatment & Diseases and treatments & 23\\
        Method & Methods of preparation of substances & 21\\
        Meta & Related to the verbatim source text, the structure of the text and external references & 38\\
        Multi-Concept & About more than one aforementioned concepts & 11\\
        Miscellaneous & Miscellaneous concepts & 24\\
        \bottomrule
    \end{tabular}
    }
    \caption{Question Categories for \ayurveda QA Dataset}
    \label{tab:catayurveda}
    \end{minipage} 
\end{table}

\section{Retrieval Augmented Generation}
\label{sec:rag}

In the RAG paradigm, the LLM is provided with additional context that consists of top-$k$ passages retrieved from the original texts. The texts of \ramayana and \bpn are obtained from GRETIL\footnote{\url{https://gretil.sub.uni-goettingen.de/}} and Sanskrit Wikisource\footnote{\url{https://sa.wikisource.org/wiki/}} respectively. The texts are pre-processed following standard procedures \citep{manning2008introduction}, namely, dividing the texts into chunks, followed by lemmatization, and then building a document store. Lemmatization would not have been necessary if retrieval frameworks such as Dense Passage Retrieval \citep{karpukhin-etal-2020-dense} or a vector space retrieval framework with SentenceBERT embeddings \citep{reimers-gurevych-2019-sentence} could be used. However, due to insufficient data in Sanskrit, such models cannot be trained now. Hence, we used BM25 retrieval and vector space retrieval with averaged FastText (AvgFT) \citep{bojanowski2017enriching} and GloVe \citep{pennington-etal-2014-glove} (AvgGV) embeddings, which are employed on lemmatized documents and queries. To achieve this, a lemmatizer for Sanskrit was built as described below.

\subsubsection*{Sanskrit Lemmatizer}

Seq2Seq transformer-based Sanskrit lemmatizer was trained from the words and their respective lemmas present in the DCS corpus \citep{hellwig2010dcs}\footnote{\url{http://www.sanskrit-linguistics.org/dcs/}}. During lemmatization, if a word in an input sentence is a compound word or involves Sandhi, the lemmatizer is expected to break the word into sub-words and generate their respective lemmas in the output. For example, if the input sentence is `\textit{haridr\={a}malaka\d{m} g\d{r}h\d{n}\={a}ti}', the corresponding lemmatized output should be `\textit{haridr\={a} \={a}malaka g\d{r}h}'. Our lemmatizer achieves a mean F1-score of 0.94 across the sentences from the held-out test set (Appx.~\ref{sec:app-impl}) calculated according to \citet{melamed-etal-2003-precision}, however with a significant standard deviation of 0.11. While the accuracy is high, future attempts for improvements should focus on minimizing the variance, which is rarely ever reported although important.

The information retrieval pipelines thus formulated can be considered novel concerning Classical Sanskrit. A known earlier attempt towards building retrieval systems in Sanskrit \citep{sahu2023building} focused on news corpora with much terminology consisting of borrowings from Hindi and even English. As a result, the lemmatizer trained on Classical Sanskrit and thereby, our entire retrieval pipeline may not be appropriate on such corpora and hence are not comparable.

The prompts for RAG are detailed  in Appx.~\ref{subsec:app-prompt-rag}.

\section{Implementation}
\label{sec:app-impl}

This appendix outlines the implementation details. All LLMs are operated through API calls using LangChain\footnote{\url{https://www.langchain.com/}}. In case of Llama-3.1, we used API provided by Fireworks AI\footnote{\url{https://fireworks.ai/}}.

The lemmatizer was implemented using HuggingFace transformers \citep{wolf-etal-2020-transformers} upon base model T5 \citep{raffel2020exploring} initiated with the model configuration of 4 layers per each encoder and decoder, 4 attention-heads, embedding of size 256, and hidden size of 1024, totaling about 100M parameters. The tokenizer trained by \citet{akavarapu-bhattacharya-2023-creation} was used\footnote{\url{https://huggingface.co/mahesh27/vedicberta-base}}. The lemmatizer was trained for 15 epochs on DCS \citep{hellwig2010dcs} data with batch size of 32, that took about 15 hours on NVIDIA RTX 2080 with 11GB graphics memory. There are total 1.04M sentences in the data, that are randomly divided into proportions $0.675:0.075:0.15$ respectively for training, validation and testing. FastText and GloVe embeddings are trained on lemmas obtained from DCS \citep{hellwig2010dcs} with embedding size 100.

\section{Supplementary Results}
\label{sec:app-add}
\begin{table*}[t]
    
    \begin{minipage}{.5\linewidth}
      
      \centering
      \resizebox{0.75\textwidth}{!}{
        \begin{tabular}{lr} \toprule
\textbf{Model}                         & \multicolumn{1}{l}{\textbf{BLEU}} \\ \midrule
Google Trans \citep{maheshwari-etal-2024-samayik} & 13.9                              \\
IndicTrans \citep{maheshwari-etal-2024-samayik}   & 13.1                              \\
gpt-4o                                 & 16.5                              \\
llama-3.1-405b-instruct                & \textbf{17.1}                    \\ \bottomrule
\end{tabular}}
        \caption*{MT (\texttt{san-eng}) on \textit{Mann ki Baat} dataset}
    \end{minipage}%
    \begin{minipage}{.5\linewidth}
      \centering
        
              \resizebox{0.85\textwidth}{!}{
        \begin{tabular}{lr} \toprule
\textbf{Model}                      & \multicolumn{1}{l}{\textbf{Macro F1 (BI)}} \\ \midrule
LatinBERT1 \citep{beersmans-etal-2023-training} & 0.54                              \\
LatinBERT2 \citep{beersmans-etal-2023-training} & 0.50                                        \\
\texttt{gpt-4o}                              & \textbf{0.55}                                       \\
\texttt{llama-3.1-405b-instruct}            & 0.36                            \\ \bottomrule
\end{tabular}}
        \caption*{NER (\texttt{lat}) on \textit{Ars Amatoria} dataset}
    \end{minipage} 
    \caption{Comparison of out of domain performances of LLMs against previously reported fine-tuned models.}
    \label{tab:results}
\end{table*}

In Table \ref{tab:results}, we compare the out-of-domain performance of our evaluated models against previously reported fine-tuned models. For MT (\texttt{san-eng}) on \textit{Mann ki Baat} dataset \citep{maheshwari-etal-2024-samayik}, open-source model \texttt{llama-3.1-405b-instruct} outperforms both Google Trans and IndicTrans, while for NER (\texttt{lat}) on Ovid's \textit{Ars Amatoria} dataset \citep{beersmans-etal-2023-training}, the performance of \texttt{gpt-4o} is better than that of fine-tuned LatinBERT variants. Although fine-tuned models yield superior results on in-domain data, our findings indicate that multilingual LLMs are superior in their zero-shot generalization.

\section{LLMs with Knowledge Graphs}
\label{subsec:kgdesc}

\begin{figure}[h!]
    \centering
    \includegraphics[width=0.8\linewidth]{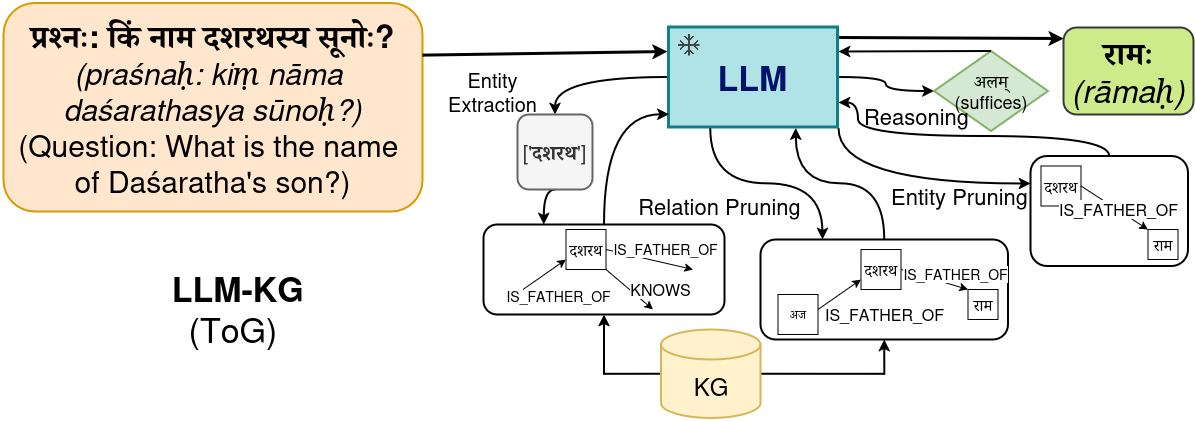}
    \caption{Overview of augmenting a LLM with a knowledge graph (KG) through Think-on-Graph (ToG) paradigm.}
    \label{fig:llmkg}
\end{figure}

Arriving at an answer by an LLM integrated with a knowledge graph (KG) through Think-on-Graph (ToG) \citep{sun2024thinkongraph} paradigm involves several prompting steps for each hop from starting entity nodes as illustrated in Fig. \ref{fig:llmkg}. Firstly, the LLM lists entities from the input questions further lemmatized by our lemmatizer previously described. The relationships from and to these entities are then extracted by traversing the KG. The LLM then lists relationships with relevance scores, which are further used to prune the relationships, retaining only the best three. Unexplored entities connected by these relationships are then known from the KG, which are similarly pruned to retain the three most relevant ones. The LLM then reasons whether these extracted paths suffice to answer the given question. If no, the cycle is repeated, i.e., it traverses a hop further up to a depth $d$. Otherwise, the LLM answers using the context from the extracted paths. 

The prompts for each step and an outline pseudo-code can be found respectively in Appx.~\ref{subsec:app-prompt-kg}, Alg.~\ref{alg:tog}. Technical terminology such as `entity', `knowledge graph', and so forth are mostly retained in English in these prompts resulting in minimal and unavoidable code-mixing. Further, the output of these prompts is often a list of elements and, hence, has to abide by a structured format.

\subsection{Knowledge Graphs}

A knowledge graph (KG) was constructed for \ramayana using two key references,
\cite{ray1984index} and \cite{rai1965kosha}. The graph was annotated with the help of two experts proficient in both Sanskrit and \ramayana. For annotation,
we used a custom deployment of \emph{Sangrahaka} \cite{terdalkar2021sangrahaka}.
The resulting knowledge graph contains \rmkgnodes\ nodes and \rmkgrelations\
relations, encompassing entities like characters of the story including humans
and divine beings, places (cities, rivers, kingdoms), and animals, and
relationships such as kinship, actions, locations, and others, highlighting
associations between the characters, natural features, and other elements from
the text.

Additionally, a work-in-progress knowledge graph for \bpn obtained
from the authors of \cite{terdalkar-etal-2023-semantic} was referenced. The KG currently
includes \bpnkgnodes nodes and \bpnkgrelations relations from 12 out of 23
chapters covering substances such as grains, vegetables, meats, metals, poisons,
dairy products, prepared substances and other miscellaneous medicinal substances.

The knowledge graphs were loaded and accessed through Neo4j\footnote{\url{https://neo4j.com/}}. Python packaage, \texttt{indic-transliteration}\footnote{\url{https://github.com/indic-transliteration/indic_transliteration_py}} is used to move among transliteration schemes of Sanskrit. The pseudo-code for our implementation of ToG \citep{sun2024thinkongraph} is given in Algorithm~\ref{alg:tog}. The sample limit $S$ is set to $15$, depth limit $D$ to $1$ and width limit $W$ to 3.

\begin{algorithm}[t]
    \caption{Outline of LLM-KG i.e., ToG \citep{sun2024thinkongraph}} \label{alg:tog}
    \begin{algorithmic}
        \Require Input: $x$
        \\ LLM prompt-chains: ExtractEntities, RelationPrune, EntityExtractPrune, Reason, Answer \\ Interface to KG: FetchRelations, FetchEntities; Depth limit: $D$; Sample limit for KG: $N$; Width limit for LLM: $W$
        \State Current Entities $E \gets$ ExtractEntities($x$)
        \State Current depth $d \gets 0$
        \State Stored Paths $P \gets$ []
        \While{$d < D$}
            \State $R \gets$ FetchRelations($E$, $N$)
            \State $R \gets$ RelationPrune($R$, $W$)
            \State $E, P \gets$ FetchEntities($E$, $R$, $P$, $N$)
            \State $E, P \gets$ EntityExtractPrune($E$, $R$, $P$, $W$)
            \If{Reason($x$, $E$, $P$)}
                \State Answer($x$, $E$, $P$)
                \State \textbf{break}
            \EndIf
            \State $d \gets d + 1$
        \EndWhile
        \If{$d = D$}
            Answer($x$, $E$, $P$)
        \EndIf
    \end{algorithmic}
\end{algorithm}

\subsection{LLM-KG Prompts}
\label{subsec:app-prompt-kg}
\subsubsection*{ExtractEntities}
\textbf{system} tvam \textit{knowledge-graph}-ta\d{h} uttar\=a\d{n}i ni\d{s}kar\d{s}yitu\d{m} pra\'sn\=at \textit{entities} vindasi ca t\=ani saha \textit{relevance-score} (0-1 madhye) samarpayasi.
\\\textit{output} ud\=ahara\d{n}am (`r\=ama\d{h}', 0.8), (`s\={i}t\=a', 0.7). tato viv\d{r}ta\d{m} m\=a kuru.
\\\textbf{human} pra\'sna\d{h}: \{\texttt{QUESTION}\} \{\texttt{CHOICES}\}

\subsubsection*{RelationPrune}
\textbf{system} tvam datta-pra\'snasya uttar\=a\d{n}i \textit{knowledge-graph}-ta\d{h} ni\d{s}kar\d{s}itu\d{m} \textit{knowledge-graph}-ta\d{h} id\=an\=i\d{m} paryanta\d{m} ni\d{s}kar\d{s}ita-sambandhebhya\d{h} ava\'sy\=ani saha \textit{relevance-score} (0-1 madhye) samarpayasi.
\\\textit{output} ud\=ahara\d{n}am (`IS\_FATHER\_OF', 0.8), (`IS\_CROSSED\_BY', 0.7), $\ldots$ . tato viv\d{r}ta\d{m} m\={a} kuru.
\\\textbf{human} pra\'sna\d{h}: \{\texttt{QUESTION}\} \{\texttt{CHOICES}\}
\\ni\d{s}kar\d{s}it\=ani sambandh\=ani: \{\texttt{RELATIONS}\}

\subsubsection*{EntityExtractPrune}
\textbf{system} tvam datta-pra\'snasya uttar\=a\d{n}i \textit{knowledge-graph}-ta\d{h} ni\d{s}kar\d{s}itu\d{m} \textit{knowledge-graph}-ta\d{h} id\=an\=i\d{m} paryanta\d{m} ni\d{s}kar\d{s}ita-sambandhebhya\d{h} ava\'sy\=ani \textit{nodes (lemmas)} saha \textit{relevance-score} (0-1 madhye) samarpayasi.
\\\textit{output} ud\=ahara\d{n}am (`r\=ama\d{h}', 0.8), (`s\={i}t\=a', 0.7). tato viv\d{r}ta\d{m} m\=a kuru.
\\\textbf{human} pra\'sna\d{h}: \{\texttt{QUESTION}\} \{\texttt{CHOICES}\}
\\ni\d{s}kar\d{s}it\=ani sambandh\=ani: \{\texttt{RELATIONS}, \texttt{ENTITIES}\}

\subsubsection*{Reason}
\textbf{system} tvam datta-pra\'snasya uttar\=a\d{n}i \textit{knowledge-graph}-ta\d{h} ni\d{s}kar\d{s}itu\d{m} \textit{knowledge-graph}-ta\d{h} id\=an\=i\d{m} paryanta\d{m} ni\d{s}kar\d{s}ita\d{m} yat-ki\~ncid pra\'snasya uttara\d{m} d\=atu\d{m} alam (1) v\=a n\=alam (0) iti vaktavyam.
\\\textit{output} 1 athav\=a 0. na anyat vadasi
\\\textbf{human} pra\'sna\d{h}: \{\texttt{QUESTION}\} \{\texttt{CHOICES}\}
\\ni\d{s}kar\d{s}itam: \{\texttt{PATHS}\}

\subsubsection*{Answer}
\textbf{system} adha\d{h} \{\texttt{TOPIC}\}-sambandhe p\d{r}\d{s}\d{t}a-pra\'snasya pratyuttara\d{m} dehi. tadapi pra\'snocitavibhaktau bhavet na tu pr\=atipadika r\=upe. tadapi ekenaiva padena yadi uttare k\=ara\d{n}am n\=apek\d{s}itam. katham kimartham ity\=adi\d{s}u ekena laghu v\=akyena uttara\d{m} dehi atra tu eka-pada-niyama\d{h} n\=asti.
\\api ca yath\=a'va\'syam adha\d{h} dattai\d{h} \textit{knowledge-graph}-ta\d{h} ni\d{s}kar\d{s}ita-vi\d{s}ayai\d{h} sah\={a}yya\d{m} g\d{r}h\=a\d{n}a. tattu sarvad\=a s\=adhu iti n\=a'sti prat\=iti\d{h}. uttaram y\=avad laghu \'sakyam t\=avat laghu bhavet.
\\\textbf{human} pra\'sna\d{h}: \{\texttt{QUESTION}\} \{\texttt{CHOICES}\}
\\ni\d{s}kar\d{s}itam: \{\texttt{PATHS}\}
\\uttaram: 

\subsection{LLM-KG Results}
\label{subsec:llmkgres}
\begin{table}[t]
    \centering
    \resizebox{0.9\linewidth}{!}{
    \begin{tabular}{l|rrrrr} \toprule
\textbf{Method} & \multicolumn{1}{l}{\textbf{\texttt{gpt-4o}}} & \multicolumn{1}{l}{\textbf{\texttt{claude-3.5-sonnet}}} & \multicolumn{1}{l}{\textbf{\texttt{gemini-1.5-pro}}} & \multicolumn{1}{l}{\textbf{\texttt{mistral-large-2}}} & \multicolumn{1}{l}{\textbf{\texttt{llama-3.1-405b-instruct}}} \\ \midrule
Closed-book       & 0.381                               & 0.242                                          & 0.148                                       & 0.333                                        & 0.346                                                \\
RAG-BM25        & 0.478                               & \textbf{0.521}                                 & 0.459                                       & 0.434                                        & 0.323                                                \\
LLM-KG          & 0.381                               & 0.254                                          & -                                           & 0.341                                        & -     \\ \bottomrule                                              
\end{tabular}}
    \caption{Exact Match (Scores) of various models (including those not part of main experiments) in Sanskrit Question-Answering task (Sanskrit Prompts) with LLM-KG paradigm compared against zero-shot and RAG-BM25 paradigms.}
    \label{tab:llmkgres}
\end{table}

The LLM-KG paradigm was evaluated exclusively using Sanskrit prompts on the two QA datasets and included additional models not part of the main experiments—namely, \texttt{claude-3.5-sonnet} \citep{anthropic2024claude35}, \texttt{gemini-1.5-pro} \citep{google2024gemini}, and \texttt{mistral-large-2} \citep{mistral2024large2}. Table \ref{tab:llmkgres} presents the results in comparison with the closed-book and RAG-BM25 paradigms. Overall, performance gains from closed-book to LLM-KG are modest and fall short of the improvements observed with RAG. This may be partly attributed to the complexity of the LLM-KG setup, which requires multi-step prompting and adherence to a structured output format. Notably, models like \texttt{gemini-1.5-pro} and \texttt{llama-3.1} frequently fail to follow this structured format, rendering them ineffective for running ToG. The strict formatting requirements may also pose challenges for other models, particularly those less adapted to Sanskrit. Interestingly, while \texttt{claude-3.5-sonnet} achieves the best results with RAG-BM25, it lags behind \texttt{gpt-4o} and \texttt{mistral-large-2} in both the closed-book and LLM-KG paradigms.

\begin{table*}[t]
    
    \begin{minipage}{.5\linewidth}
      
      \centering
      \resizebox{0.95\textwidth}{!}{
        \begin{tabular}{l|rrr} \toprule
\textbf{Method} & \textbf{\texttt{gpt-4o}} & \textbf{\texttt{claude-3.5-sonnet}} & \textbf{\texttt{mistral-large-2}} \\ \midrule
closed-book     & 0.32            & 0.21                       & 0.25                     \\
LLM-KG          & 0.34            & 0.34                       & 0.35    \\ \bottomrule                
\end{tabular}
}
        \caption*{(a)}
    \end{minipage}%
    \begin{minipage}{.5\linewidth}
      \centering
        
              \resizebox{0.95\textwidth}{!}{
       \begin{tabular}{l|rrr} \toprule
\textbf{Method} & \textbf{\texttt{gpt-4o}} & \textbf{\texttt{claude-3.5-sonnet}} & \textbf{\texttt{mistral-large-2}} \\ \midrule
closed-book     & 0.40            & 0.25                       & 0.36                     \\
LLM-KG          & 0.39            & 0.23                       & 0.34            \\ \bottomrule                                              
\end{tabular}
}
        \caption*{(b)}
    \end{minipage} 
     \caption{Comparison of Exact Match (EM) scores between closed-book and LLM-KG paradigms for selected questions when the answer (a) can likely be inferred from KG and (b) cannot be inferred from KG.}
    \label{tab:llmkgrel}
\end{table*}

Table~\ref{tab:llmkgrel} presents a breakdown of performance based on whether the question topics are covered in the current KG—specifically, the \emph{kingdoms} category (27 questions) in the \ramayana dataset and the annotated chapters (299 questions) in \bpn. For these subsets, which are likely answerable from the KG, LLM-KG shows clear improvements over the closed-book setting, indicating that access to a near-complete KG can significantly enhance performance. In contrast, for questions outside these categories or chapters, no such improvement is observed, reinforcing the hypothesis that KG completeness is crucial for the effectiveness of LLM-KG. Determining domains where knowledge graphs may outperform or be more appropriate than RAG remains an open question for future research.

\section{Categories for Named Entity Recognition}
\label{app:nertypes}
The categories for NER in Sanskrit, Ancient Greek, and Latin, along with their rough translation and brief explanations, wherever applicable, are provided here.

\twocolumn
\begin{table}[h!]
\centering
\resizebox{0.55\textwidth}{!}{
\begin{tabular}{lll} \toprule
\textbf{Entity Type} & \textbf{Translation} & \textbf{Description} \\ \midrule
Manuṣya         & Human                    & A mortal human being                               \\
Deva            & Deity                    & Divine celestial being; god or goddess             \\
Gandharva       & $\sim$                   & Heavenly musician in the service of the gods       \\
Apsaras         & $\sim$                   & Beautiful female spirits known for dance and charm \\
Yakṣa           & $\sim$                   & Guardian spirit of natural treasures.              \\
Kinnara         & $\sim$                   & Certain Semi-divine beings                         \\
Rākṣasa         & $\sim$                   & Malevolent being                                   \\
Asura           & Anti-god                 & Powerful beings opposed to the gods                \\
Vānara          & Monkey-being             & Monkey-like humanoid                               \\
Bhallūka        & Bear-being               & Bear or Bear-like humanoid                         \\
Gṛdhra          & Vulture-being            & Vulture-like being                                 \\
\d{R}k\d{s}a            & Bear-being               & Bear-like humanoid                                 \\
Garuḍa          & Eagle-being              & Eagle-like being                                   \\
Nāga            & Serpent-being            & Semi-divine serpent race                           \\
Svarga          & Heaven                   & Abode of the gods                                  \\
Naraka          & Hell                     & Realm of punishment after death                    \\
Nadī            & River                    & Flowing body of freshwater                         \\
Sāgara          & Sea                      & Vast saltwater body                                \\
Sarovara        & Lake                     & Large inland water body                            \\
Kūpa            & Well                     & Man-made water source                              \\
Tīra            & Riverbank                & Edge or shore of a river                           \\
Dvīpa           & Island                   & Land surrounded by water                           \\
Parvata         & Mountain                 & Large natural elevation of earth                   \\
Nagara          & City                     & Urban settlement or metropolis                     \\
Tīrtha          & Sacred Place             & Holy pilgrimage spot, often near water             \\
Grāma           & Village                  & Small rural settlement                             \\
Rājya           & Kingdom                  & Territory ruled by a king                          \\
Vana            & Forest                   & Dense growth of trees; wilderness                  \\
Udyāna          & Garden                   & Cultivated green space                             \\
Marubhūmi       & Desert                   & Dry, arid region                                   \\
Prāsāda         & Palace                   & Royal residence                                    \\
Mandira         & Temple                   & Sacred structure for worship                       \\
Āśrama          & Hermitage                & Secluded place for spiritual practice              \\
Gṛha            & House                    & Dwelling or home                                   \\
Kuṭīra          & Hut                      & Small and simple shelter                           \\
Guhā            & Cave                     & Natural underground chamber                        \\
Mārga           & Road                     & Pathway or route                                   \\
Ratha           & Chariot                  & Two- or four-wheeled ancient vehicle               \\
Vimāna          & Airborne Vehicle         & Flying chariot or aircraft                         \\
Khadga          & Sword                    & Bladed weapon                                      \\
Dhanus          & Bow                      & Weapon for shooting arrows                         \\
Bāṇa            & Arrow                    & Projectile shot from a bow                         \\
Cakra           & Discus                   & Spinning circular weapon                           \\
Gadā            & Mace                     & Blunt weapon, often spiked                         \\
Tomara          & Javelin                  & Thrown spear or missile                            \\
Śūla            & Spear                    & Long-shafted piercing weapon                       \\
Kavaca          & Shield                   & Defensive armor piece                              \\
Kañcuka         & Armor                    & Protective body gear                               \\
Paraśu          & Axe                      & Bladed tool/weapon                                 \\
Astra           & Divine Weapon            & Supernatural weapon, often invoked                 \\
Ābharaṇa        & Ornament                 & Decorative jewelry                                 \\
Śaṅkha          & Conch                    & Sacred spiral shell                                \\
Vādya           & Musical Instrument       & Instrument used in music                           \\
Nāṇa            & Currency                 & Form of money or coin                              \\
Kula            & Clan                     & Extended family or lineage                         \\
Jāti            & Species                  & Species/Socio-economical Group                     \\
Gaṇa            & Tribe / Group            & Assembly or community                              \\
\d{R}tu             & Season                   & Climatic period of the year                        \\
Sa\d{m}vatsara      & Year                     & Vedic year cycle                                   \\
Māsa            & Month                    & Lunar or solar month                               \\
Tithi           & Lunar Day                & Phase in the moon's waxing/waning                  \\
Pakṣa           & Fortnight                & Half of a lunar month                              \\
Ayana           & Solstice Cycle           & Six-month movement of the sun                      \\
Yuga            & Epoch                    & Cosmic age or era                                  \\
Yoga            & Astronomical Combination & Planetary conjunction                              \\
Karaṇa          & Half of Tithi            & Subdivision of a lunar day                         \\
Muhūrta         & Moment / Auspicious Time & Small unit of time (about 48 minutes)              \\
Lagna           & Ascendant                & Zodiac rising at time of birth                     \\
Graha           & Planet                   & Celestial influencer                               \\
Nakṣatra        & Lunar Mansion            & One of 27 lunar constellations                     \\
Rāśi            & Zodiac Sign              & Segment of the zodiac                              \\
Dhuma-ketu      & Comet                    & Celestial object with a tail                       \\
Utsava          & Festival                 & Celebratory event                                  \\
Pūjā            & Worship                  & Ritual offering and prayer                         \\
Yajña           & Vedic Sacrifice          & Sacred fire ritual                                 \\
Upacāra         & Ritual Offering          & Ceremonial gesture or item                         \\
Sa\d{m}skāra        & Life-Cycle Rite          & Hindu ritual of life transition                    \\
Aniścita        & Undecided                & Something that is not yet determined               \\
Vṛkṣa           & Tree                     & Large woody plant                                  \\
Guccha          & Shrub                    & Small bushy plant                                  \\
Lata            & Vine                     & Climbing or trailing plant                         \\
Puṣpa           & Flower                   & Blossom of a plant                                 \\
Phala           & Fruit                    & Edible plant product                               \\
Patra           & Leaf                     & Green foliage part                                 \\
Stambha         & Stem                     & Main structural plant part                         \\
Tvak            & Bark                     & Outer layer of tree                                \\
Mūla            & Root                     & Underground part of plant                          \\
Pakṣī           & Bird                     & Feathered flying animal                            \\
Sarpa           & Snake                    & Legless reptile                \\ \bottomrule                 
\end{tabular}
}
\caption{Entity types occuring in Sanskrit NER}
\end{table}

\begin{table}[h!]
\centering
\resizebox{0.40\textwidth}{!}{
\begin{tabular}{ll} \toprule
\textbf{Entity Type} & \textbf{Description}                              \\ \midrule
NORP                 & Ethnic groups, demonyms, schools                  \\
ORG                  & Organizations                                     \\
GOD                  & Supernatural beings                               \\
LANGUAGE             & Languages and dialects                            \\
LOC                  & Cities, empires, rivers, mountains, and so forth. \\
PERSON               & Individual persons                               \\ \bottomrule
\end{tabular}
}
\caption{Entity types occuring in Ancient Greek NER \citep{myerston-nereus-2025}. The types without descriptions---EVENT and WORK---have very few occurances in the dataset.}
\end{table}

\begin{table}[h!]
\centering
\resizebox{0.30\textwidth}{!}{
\begin{tabular}{ll} \toprule
\textbf{Entity Type} & \textbf{Description}        \\ \midrule
PER                  & Person                      \\
LOC                  & Locations, places           \\
GRP                  & Other groups such as tribes \\ \bottomrule
\end{tabular}
}
\caption{Entity types occuring in Latin NER are quite standard types.}
\end{table}

\end{document}